\documentclass[a4paper,fleqn]{cas-dc}

\usepackage[numbers,sort&compress]{natbib}
\usepackage[commandnameprefix=always]{changes}
\usepackage{subcaption}
\usepackage{xcolor}



\usepackage{tabularx} 
\usepackage{amsthm}

\usepackage{array}
\usepackage{algorithm}
\usepackage{algpseudocode}
\usepackage{booktabs}
\usepackage{threeparttable}
\usepackage[normalem]{ulem}
\theoremstyle{remark}          
\newtheorem{remark}{Remark}    
\hypersetup{
  colorlinks=false,        
 pdfborder={0 0 1},       
            citebordercolor={0 1 0}, 
            linkbordercolor={1 0 0}, 
            urlbordercolor={0 1 1}   
}

\begin{document}
\let\WriteBookmarks\relax
\def\floatpagepagefraction{1}
\def\textpagefraction{.001}

\shorttitle{Vision–Proprioception Fusion with Mamba2 in End-to-End Reinforcement Learning for Motion Control}

\shortauthors{Xiaowen Tao, Yinuo Wang, and Jinzhao Zhou}

\title [mode = title]{\textcolor{black}{Vision–Proprioception Fusion with Mamba2 in End-to-End Reinforcement Learning for Motion Control}}             


\author[1]{Xiaowen Tao}[style=chinese]
\ead{taox@tcd.ie}
\credit{Conceptualization, Methodology, Formal analysis, Writing- Original Draft, Writing- Review \& Editing, Supervision}

\author[2]{Yinuo Wang}[style=chinese]
\cormark[1]
\ead{allen.wang@woven.toyota}
\credit{Conceptualization, Methodology, Software, Validation, Investigation, Resources, Visualization, Writing- Original Draft, Writing- Review \& Editing}

\author[3]{Jinzhao Zhou}[style=chinese]
\ead{jinzhao.zhou@uts.edu.au}
\credit{Writing- Review \& Editing}

\affiliation[1]{organization={School of Computer Science and Statistics, Trinity College Dublin},
    city={Dublin},
    country={Ireland}}

\affiliation[2]{organization={Woven by Toyota},
    city={Palo Alto},
    state={CA},
    country={USA}}

\affiliation[3]{organization={Faculty of Engineering and Information Technology, University of Technology Sydney},
    city={Sydney},
    country={Australia}}

\cortext[cor1]{Corresponding author}

\begin{abstract}
End-to-end reinforcement learning (RL) for motion control trains policies directly from sensor inputs to motor commands, enabling unified controllers for different robots and tasks. However, most existing methods are either \emph{blind} (proprioception-only) or rely on fusion backbones with unfavorable compute–memory trade-offs. Recurrent controllers struggle with long-horizon credit assignment, and Transformer-based fusion incurs quadratic cost in token length, limiting temporal and spatial context. We present a vision-driven cross-modal RL framework built on \emph{SSD-Mamba2}, a selective state-space backbone that applies \emph{state-space duality (SSD)} to enable both recurrent and convolutional scanning with hardware-aware streaming and near-linear scaling. Proprioceptive states and exteroceptive observations (e.g., depth tokens) are encoded into compact tokens and fused by stacked SSD-Mamba2 layers. The selective state-space updates retain long-range dependencies with markedly lower latency and memory use than quadratic self-attention, enabling longer look-ahead, higher token resolution, and stable training under limited compute. Policies are trained end-to-end under curricula that randomize terrain and appearance and progressively increase scene complexity. A compact, state-centric reward balances task progress, energy efficiency, and safety. Across diverse motion-control scenarios, our approach consistently surpasses strong state-of-the-art baselines in return, safety (collisions and falls), and sample efficiency, while converging faster at the same compute budget. These results suggest that SSD-Mamba2 provides a practical fusion backbone for resource-constrained robotic and autonomous systems in engineering informatics applications.

\end{abstract}


\begin{highlights}
\item \textbf{Establishing SSD-Mamba2 as a backbone for RL motion control.} We present the first study to leverage the SSD–based Mamba2 architecture in reinforcement learning for end-to-end motion control, demonstrating its suitability as an efficient and scalable sequence model in this domain.

\item \textbf{Demonstrating effective cross-modal fusion.} We show that SSD-Mamba2 provides an efficient and expressive backbone for integrating proprioceptive states with depth-image tokens. Its near-linear scaling, stable recurrent updates, and ability to capture long-range dependencies enable foresightful and robust control, outperforming Transformer-based and proprioception-only approaches.

\item \textbf{Validating end-to-end training with PPO.} We integrate SSD-Mamba2 into a PPO pipeline, complemented by domain randomization, obstacle-density curricula, and a compact state-centric reward. This design yields faster convergence, higher training efficiency, and more stable performance than established baselines.

\item \textbf{Comprehensive empirical evaluation.} Through extensive experiments across diverse motion-control tasks, we demonstrate that SSD-Mamba2 improves performance, reduces collisions and falls, and generalizes better than state-of-the-art methods, supporting its potential for safety-critical robotic applications.
\end{highlights}


\maketitle

\section{Introduction}
\label{sec:first}
Quadrupedal robots are increasingly deployed in domains where safe and reliable mobility is critical, including inspection, disaster response, and exploration of unstructured terrains \citep{imambi2021pytorch}. Unlike wheeled platforms, they can traverse rubble, stairs, and deformable substrates, but their control remains a longstanding challenge: locomotion must be robust under diverse dynamics, foresightful to anticipate hazards, and computationally efficient for deployment on embedded hardware \citep{carpentier2021recent}.

Deep reinforcement learning (DRL) offers a promising route toward such capabilities by training policies end-to-end from interaction. While DRL has enabled agile behaviors across terrains and disturbances, most prior controllers fall short in one of two ways. Proprioception-only (``blind'') agents can maintain stability but lack foresight, reacting only after contact with obstacles \citep{xie2022glide}. In contrast, cross-modal approaches that fuse vision with proprioception provide anticipatory information but rely on backbones with unfavorable compute–memory trade-offs \citep{han2025multimodal}. Recurrent models struggle to capture long-horizon dependencies \citep{li2024rapid}, hierarchical schemes complicate optimization \citep{jain2019hierarchical}, and Transformers incur quadratic costs in token length, limiting their ability to scale \citep{yang2021learning}. These limitations raise a broader question: \emph{Can recent advances in sequence modeling unlock practical end-to-end RL for motion control?} 

\textcolor{black}{Recent state-space models (SSMs) provide an appealing alternative by combining recurrent structure with efficient sequence processing, but different SSM variants exhibit distinct trade-offs when applied to RL. Structured long-convolution models such as S4 \citep{gu2021efficiently} rely on explicit continuous-to-discrete parameterizations and long convolution kernels, often involving additional design choices such as kernel truncation and fast Fourier transform--based implementations \cite{beaudoin2002accurate}. While effective for offline or supervised sequence modeling, these design elements can increase hyperparameter sensitivity and tuning complexity in RL settings with long rollouts and non-stationary data distributions. Selective SSMs such as Mamba \citep{gu2023mamba} address some of these challenges through input-dependent state updates and efficient scan operations. However, in Mamba, the link between recurrent state updates and their scan-based implementation remains implicit. In contrast, SSD-Mamba2 \citep{dao2024transformersssmsgeneralizedmodels} is built on an explicit state-space duality (SSD) formulation, in which recurrent evolution and block-wise parallel scanning are derived from the same state-space representation. This explicit mapping enables a unified and implementation-efficient realization of recurrent computation, reducing design complexity and improving practical robustness \cite{ibrahim2025survey}.} 

\textcolor{black}{
These properties make SSD-Mamba2 well-suited for real-time robotic motion control. Its hardware-aware recurrent scanning achieves near-linear time and memory complexity while supporting streaming execution and long-range temporal modeling. This balance between recurrent stability and computational efficiency is critical for quadrupedal locomotion, where policies must continuously fuse proprioceptive feedback and depth observations under strict real-time and resource constraints.}

In this paper, we introduce a vision-driven cross-modal RL framework built on \emph{SSD-Mamba2}. Proprioceptive states are embedded by a lightweight multilayer perceptron (MLP) \citep{tolstikhin2021mlpmixerallmlparchitecturevision}, depth images are patchified into spatial tokens by a compact convolutional neural network (CNN) \citep{oshea2015introductionconvolutionalneuralnetworks}, and the resulting token sequence is fused by stacked SSD-Mamba2 layers. Policies and value functions are then trained end-to-end with Proximal Policy Optimization (PPO) \citep{schulman2017proximal} under domain randomization and an obstacle-density curriculum. A compact state-centric reward further encourages progress, energy efficiency, and safety \citep{yang2022learning}.

\textbf{The main contributions of this paper are summarized as follows:}

\begin{enumerate}[1)]
\item \textbf{Establishing SSD-Mamba2 as a backbone for RL motion control.} We present the first work to leverage the SSD–based Mamba2 architecture in RL for end-to-end motion control, demonstrating its suitability as an efficient and scalable sequence model in this domain.

\item \textbf{Demonstrating effective cross-modal fusion.} We show that SSD-Mamba2 provides an efficient and expressive backbone for integrating proprioceptive states with depth-image tokens. Its near-linear scaling, stable recurrent updates, and ability to capture long-range dependencies enable foresightful and robust control, outperforming Transformer-based and proprioception-only approaches.

\item \textbf{Validating end-to-end training with PPO.} We integrate SSD-Mamba2 into a PPO pipeline, complemented by domain randomization, obstacle-density curricula, and a compact state-centric reward. This design yields faster convergence, higher training efficiency, and more stable performance than established baselines.

\item \textbf{Comprehensive empirical evaluation.} Through extensive experiments across diverse motion-control tasks, we demonstrate that SSD-Mamba2 improves performance, reduces collisions and falls, and generalizes better than state-of-the-art methods, supporting its potential for safety-critical robotic applications.
\end{enumerate}

The remainder of this paper is organized as follows. Section~\ref{sec:second} reviews related work on blind and cross-modal locomotion. Section~\ref{sec:third} presents the methodology, including the problem formulation, the SSD-Mamba2 fusion backbone, and PPO optimization. Section~\ref{sec:fourth} describes the experimental setup and implementation details. Section~\ref{sec:results} presents the experimental results and ablation studies. Section~\ref{sec:fifth} concludes the paper and discusses directions for future research.

\section{Related Work}
\label{sec:second}

\subsection{Quadrupedal Locomotion Control}
Research on quadrupedal locomotion has traditionally followed two paradigms: model-based control and learning-based methods. Early rule-based approaches relied on template dynamics, heuristics, and hand-engineered gait patterns for balance and foot placement \citep{miura1984dynamic,liu2014planning,habu2018simple,bledt2018cheetah}, providing insights into leg coordination but limited adaptability to diverse environments. Later, model-based pipelines advanced to constrained optimal control, including trajectory optimization over centroidal or full-body dynamics and model predictive control (MPC) with explicit contact constraints and task hierarchies \citep{grandia2019feedback,yao2021hierarchy,amatucci2024accelerating,elobaid2025adaptive}. These methods achieved robust disturbance rejection and stable locomotion in structured settings, with examples such as MPC-based whole-body controllers for stair climbing and rapid trotting \citep{di2018dynamic,ding2019real,carius2019trajectory}. However, such controllers require accurate models of dynamics, ground contact, and friction, along with significant manual cost tuning. This reliance limits their scalability to unstructured or rapidly changing terrains, such as rubble or deformable substrates.

By contrast, learning-based methods aim to reduce modeling effort by training adaptive policies directly from interaction \cite{wang2026locomamba}. Model-free DRL in particular has shown strong potential for quadrupeds in diverse scenarios \citep{li2021reinforcement,zhang2021terrain,bussola2025guided}. With large-scale domain randomization, policies trained in simulation have been successfully transferred to hardware \citep{tan2018sim,hwangbo2019learning}. Recent work has further improved adaptability, such as rapid motor adaptation \citep{kumar2021rma,wang2025quadkan}, which augments a base policy with an adaptation module to handle payload changes and novel terrains online. DRL policies have also demonstrated robustness under water, mud, and snow conditions \citep{lee2020learning,xie2021dynamics,wang2025humam}. Despite these advances, the majority of policies remain proprioception-only. Blind controllers can recover from disturbances but lack foresight and react only after contact events \citep{xie2022glide}, which limits proactive obstacle avoidance and foothold planning. This motivates incorporating exteroceptive sensing such as vision to endow quadrupeds with anticipatory capability.

\subsection{Vision-Driven Reinforcement Learning}
Vision has emerged as a powerful complement to proprioception by providing spatial foresight. Visual information enables robots to perceive obstacles before contact, anticipate irregular terrain, and plan footholds or body trajectories proactively. Several studies have incorporated visual input directly into DRL pipelines. For instance, \citet{yu2021visual} showed that depth-augmented RL policies could traverse rough terrain with greater robustness than proprioceptive baselines. \citet{duan2024learning} combined heightmap observations with a depth-to-heightmap predictor, enabling sim-to-real transfer on hardware. ViTAL \citep{fahmi2022vital} used vision for terrain-aware foothold selection and pose adaptation, improving safety on irregular surfaces. Other works explored semantic segmentation or heightmap predictors to enhance terrain awareness \citep{han2025multimodal}. Hierarchical designs \citep{jain2019hierarchical} further decomposed vision-based locomotion into high-level planning and low-level control.

However, fusion strategies for vision and proprioception present trade-offs. Simple feature concatenation underuses spatial structure and temporal dependencies, leading to limited generalization. Hierarchical formulations complicate optimization and can propagate errors across levels. Transformer-based fusion improves foresight but incurs quadratic time and memory scaling in token length, limiting input resolution and training efficiency \citep{yang2021learning,singh2022reinforcement}. As a result, despite their promise, many vision-driven RL frameworks remain difficult to scale or deploy efficiently. This motivates the search for backbones that retain foresight while being more computationally tractable.

\subsection{Sequence Models for Cross-Modal Fusion}
A central challenge in vision-driven RL is the design of effective cross-modal fusion backbones. Recurrent neural network (RNN)–based approaches, such as long short-term memory (LSTM) and gated recurrent unit (GRU) architectures, have been used to integrate visual and proprioceptive information for quadrupedal locomotion \citep{xiao2024egocentric,lai2024world}. These methods capture temporal context but suffer from vanishing gradients and limited long-horizon memory, especially in extended locomotion sequences. Transformer-based models have recently become popular for cross-modal fusion \citep{yang2021learning}, offering strong representational power and long-range context modeling. However, Transformers scale quadratically with sequence length, which constrains spatial resolution, temporal context, and training efficiency. This makes them costly for real-time quadrupedal control, where compute and memory budgets are limited.

State-space models (SSMs) offer an alternative by updating compact recurrent states with near-linear complexity \citep{wang2026locomamba}. The S4 model \citep{gu2021efficiently} first demonstrated efficient long-range dependency modeling on long-sequence benchmarks. Building on this, Mamba \citep{gu2023mamba} introduced selective state updates with hardware-aware parallel scanning, achieving linear-time sequence modeling with strong throughput and memory efficiency. Variants such as AlignMamba \citep{li2025alignmamba}, FusionMamba \citep{xie2024fusionmamba}, and DepMamba \citep{ye2025depmamba} have extended this paradigm to multimodal tasks, showing improvements in alignment, fusion, and cross-modal consistency. More recently, Mamba2 advanced this line of work through the SSD formulation, which further enhances stability and expressivity while retaining near-linear scaling \citep{dao2024transformersssmsgeneralizedmodels}. SSD-Mamba2 combines the efficiency of selective scanning with dual state representations, enabling more reliable modeling of long-horizon dependencies. 

To the best of our knowledge, this is the first work to use cross-modal SSD-Mamba2 for quadrupedal locomotion.

\section{Methodology}
\label{sec:third}

\subsection{Overview}
\begin{figure*}
\centering
\includegraphics[width=1.0\linewidth]{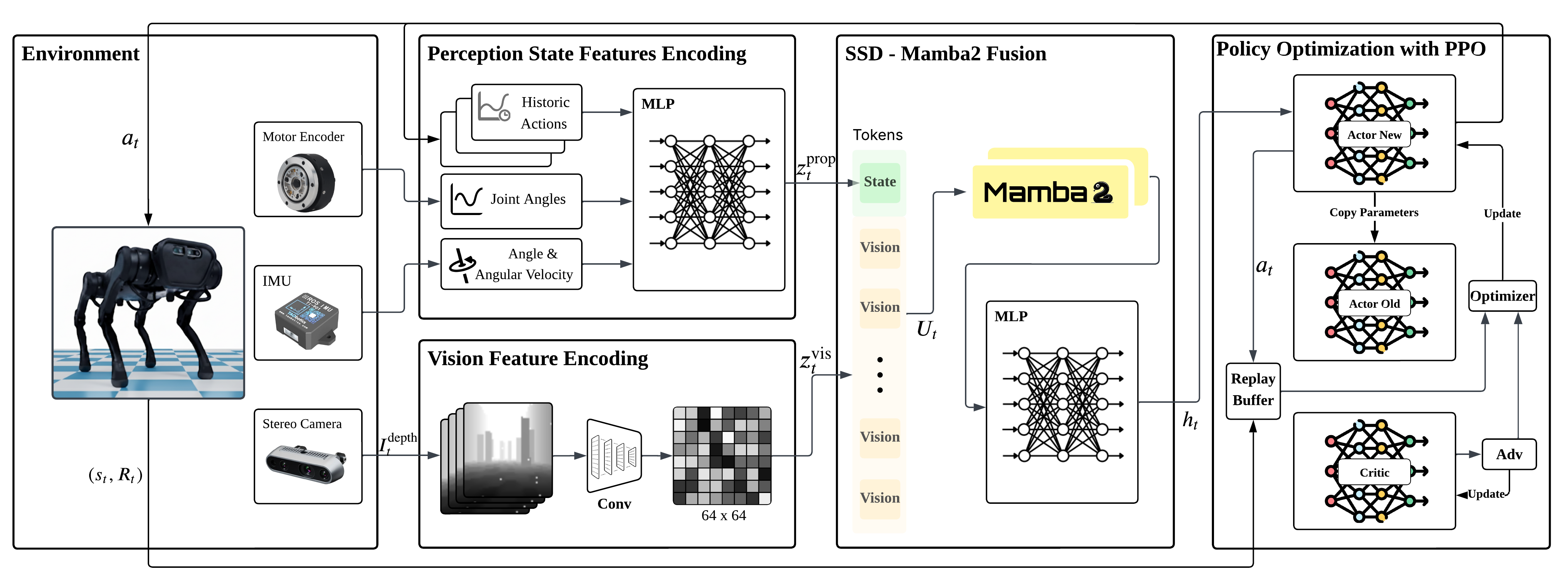}
\caption{\textcolor{black}{Overview of the proposed architecture. Proprioception and depth are encoded into tokens by MLP/CNN, fused by a SSD-Mamba2 backbone, and optimized with PPO.}}
\label{fig:overall_archi}
\end{figure*}

We propose an end-to-end reinforcement learning framework that employs the SSD–based Mamba2 architecture as a fusion backbone for quadrupedal motion control. The overall pipeline is illustrated in Fig.~\ref{fig:overall_archi}. At each time step, the agent receives both proprioceptive and visual observations. Proprioceptive states are embedded by a lightweight MLP, while depth images are patchified into spatial tokens by a compact CNN. These tokens are fused by stacked SSD-Mamba2 layers (Fig.~\ref{fig:mamba}) through hardware-aware selective scanning, which achieves near-linear time and memory scaling. This design preserves long-horizon dependencies while reducing latency and memory usage compared with quadratic self-attention. The recurrent formulation also accommodates variable token lengths and resolutions, enabling a broader perceptual context, while input-gated, exponentially decaying dynamics provide a stabilizing inductive bias. The fused representation is consumed by policy and value heads, which are optimized end-to-end using PPO. Training is further enhanced by domain randomization and an obstacle-density curriculum, which expand environmental diversity and progressively increase task difficulty. A compact state-centric reward encourages balanced policies that make forward progress efficiently and safely.

\subsection{Problem Formulation}

1) \textbf{Markov Decision Process:}  
The motion-control task is formulated as a Markov Decision Process (MDP) \citep{singh2022reinforcement}
$\mathcal{M}=(\mathcal{S}, \mathcal{A}, \mathcal{P}, r, \gamma)$, 
where $\mathcal{S}$ is the state space, $\mathcal{A}$ is the action space, 
$\mathcal{P}(s_{t+1}\mid s_t,a_t)$ is the transition distribution defined by the physics simulator, 
and $r(s_t,a_t)$ is the reward function. 
The goal is to learn a policy $\pi_\theta(a_t \mid s_t)$ that maximizes the expected discounted return:
\begin{equation}
J(\pi) \;=\; \mathbb{E}_{\pi}\!\left[ \sum_{t=0}^{T-1} \gamma^t r(s_t,a_t) \right],
\end{equation}
with discount factor $\gamma \in (0,1)$.

2) \textbf{Observation space:}  
At each time step $t$, the agent receives a multimodal observation:
\begin{equation}
s_t = \Big\{\, s^{\text{prop}}_t,\; I^{\text{depth}}_{t-3:t} \,\Big\},
\end{equation}
where $s^{\text{prop}}_t \in \mathbb{R}^{93}$ is a proprioceptive vector containing IMU readings, local joint rotations, 
and the actions executed in the previous three steps.  
The term $I^{\text{depth}}_{t-3:t} = \{I^{\text{depth}}_{t-3}, I^{\text{depth}}_{t-2}, I^{\text{depth}}_{t-1}, I^{\text{depth}}_{t}\}$ 
denotes a stack of four recent depth frames, each with resolution $64 \times 64$.

3) \textbf{Action space:}  
The action $a_t \in \mathbb{R}^{12}$ specifies the desired joint position targets for the robot’s 12 actuated joints. 
A proportional–derivative (PD) controller \citep{tan2009computation} converts these position targets into torque commands applied to the actuators.

4) \textbf{Reward function:}  
The reward encourages forward progress while discouraging unsafe or inefficient behaviors. 
It is defined as

\begin{equation}
R_t = \alpha_{\text{fwd}} R^{\text{fwd}}_t 
    + \alpha_{\text{energy}} R^{\text{energy}}_t
    + \alpha_{\text{alive}} R^{\text{alive}}_t
    + K_t \cdot R^{\text{sphere}}_t,
\end{equation}
with coefficients $\alpha_{\text{fwd}}{=}1$, $\alpha_{\text{energy}}{=}0.005$, and $\alpha_{\text{alive}}{=}0.1$.  

The forward term rewards task-aligned locomotion,
\begin{equation}
R^{\text{fwd}}_t = \langle v_t,\ e_x \rangle,
\end{equation}
where $v_t$ is the base linear velocity and $e_x$ is the unit vector along the desired direction.  
The energy term penalizes excessive control effort,
\begin{equation}
R^{\text{energy}}_t = -\,\|\tau_t\|_2^2,
\end{equation}
where $\tau_t$ denotes realized joint torques.  

The alive term $R^{\text{alive}}_t$ encourages the agent to remain operational. 
It provides a constant reward of $1.0$ at each time step until termination. 
Termination is triggered by dangerous behaviors such as falling down or colliding irrecoverably with obstacles.

The sphere collection reward $R^{\text{sphere}}_t$ (whenever applicable) is added whenever the agent collects a sphere, where $K_t$ denotes the number of 
spheres collected at the current time step.

This compact, state-centric reward balances task performance, efficiency, and safety, facilitating stable on-policy learning.

\subsection{SSD-Mamba2 Fusion Backbone}
\label{sec:ssd-mamba2}

{\setlength{\abovedisplayskip}{3pt}%
 \setlength{\belowdisplayskip}{3pt}%

The SSD-Mamba2 backbone integrates proprioceptive and visual observations into a unified representation for control. 
It extends SSMs with SSD, which equips the model with both a recurrent and a convolutional formulation, ensuring efficient long-horizon modeling with near-linear complexity. 
The backbone can be described in four components:

\begin{figure}
\centering
\includegraphics[width=\linewidth]{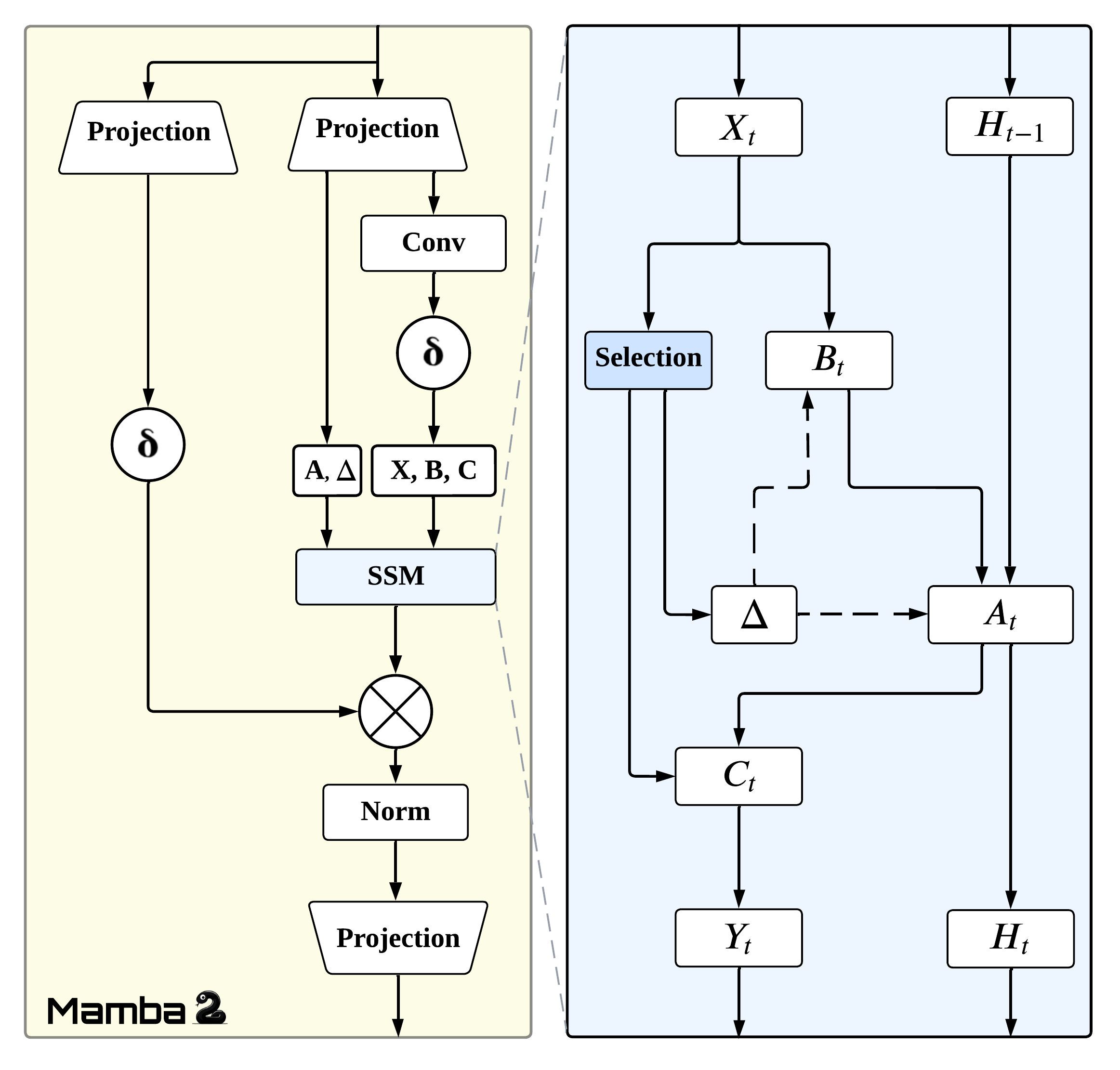}
\caption{Schematic of a SSD-Mamba2 Layer.}
\label{fig:mamba}
\end{figure}

1) \textbf{Tokenization of multimodal inputs:}  
At each time step $t$, the proprioceptive state $s_t^{\text{prop}} \in \mathbb{R}^{D_p}$ is mapped into a token
\begin{equation}
z_t^{\text{prop}} = W_p f_{\text{MLP}}(s_t^{\text{prop}}), \quad z_t^{\text{prop}} \in \mathbb{R}^d,
\end{equation}
while the depth image $I_t^{\text{depth}} \in \mathbb{R}^{H \times W}$ is patchified into $N$ tokens
\begin{equation}
z_t^{\text{vis}} = W_v f_{\text{CNN}}(I_t^{\text{depth}}), \quad z_t^{\text{vis}} \in \mathbb{R}^{N \times d}.
\end{equation}
The concatenated sequence is
\begin{equation}
U_t = \big[\, z_t^{\text{prop}};\; z_t^{\text{vis}} \,\big] \in \mathbb{R}^{(1+N)\times d}.
\end{equation}

2) \textbf{Selective state-space update:}  
For token $u_{t,k}$, each SSD-Mamba2 layer maintains a recurrent state $x_{t,k}$ and updates it with input-gated dynamics:
\begin{equation}
x_{t,k+1} = \sigma(W_A u_{t,k})\,x_{t,k} + \sigma(W_B u_{t,k})\,u_{t,k},
\end{equation}
\vspace{-1\baselineskip}
\begin{equation}
y_{t,k} = \sigma(W_C u_{t,k})\,x_{t,k} + \sigma(W_D u_{t,k})\,u_{t,k},
\end{equation}
where $\sigma(\cdot)$ denotes elementwise gating functions. This recurrent formulation yields an implicit convolutional view,
\begin{equation}
y_{t,k} = \sum_{i=0}^k K_{k,i}(u_{t,\cdot})\,u_{t,k-i},
\end{equation}
with exponentially decaying coefficients $K_{k,i}$ that stabilize long-horizon modeling.

3) \textbf{Stacked backbone:}  
With $L_m$ stacked layers, hidden states are refined with residual connections and normalization:
\begin{equation}
H_t^{(\ell+1)} = \mathrm{LN}\!\big(Y_t^{(\ell)} + H_t^{(\ell)}\big), 
\quad \ell=0,\ldots,L_m-1.
\end{equation}

4) \textbf{Output representation:}  
The final representation aggregates spatial tokens and fuses them with the proprioceptive token:
\begin{equation}
\bar{y}_t^{\text{vis}} = \tfrac{1}{N}\sum_{i=1}^N y_{t,i}^{\text{vis}}, \quad
h_t = f_{\text{head}}\!\left([\,y_t^{\text{prop}};\,\bar{y}_t^{\text{vis}}]\right).
\end{equation}
}

\begin{remark}
For quadrupedal motion control, SSD-Mamba2 confers three practical benefits. 
First, its near-linear scaling in tokens reduces the inference latency and memory footprint, which is critical for onboard deployment under real-time constraints. Second, the state-space duality formulation enables stable long-horizon dependency modeling, allowing the agent to anticipate terrain changes and obstacles beyond immediate sensor readings. 
Third, the input-gated dynamics provide robustness to sensor noise and varying image resolutions, mitigating performance degradation when proprioceptive or visual signals are imperfect. Together, these properties make SSD-Mamba2 a strong fusion backbone for safety-critical locomotion, where foresight, efficiency, and robustness are indispensable.
\end{remark}

\subsection{Policy Optimization with PPO}

PPO is a widely used model-free reinforcement learning algorithm for continuous control due to its stability and simplicity. 
Instead of performing large, potentially destabilizing policy updates, PPO restricts the update step with a clipping mechanism, ensuring stable and efficient training \citep{schulman2017proximal}. The objective of PPO is defined as:
\begin{equation}
\pi^\ast = \arg\max_\pi \;
\mathbb{E}_t \Big[ \min\big(\rho_t(\theta) \hat{A}_t,\;
\text{clip}(\rho_t(\theta),\, 1-\epsilon,\, 1+\epsilon)\,\hat{A}_t \big)\Big],
\end{equation}
where $\pi$ denotes the policy, $\hat{A}_t$ is the estimated advantage at time $t$, 
and $\rho_t(\theta)$ is the importance sampling ratio:
\begin{equation}
\rho_t(\theta) = \frac{\pi_\theta(a_t \mid s_t)}{\pi_{\theta_{\text{old}}}(a_t \mid s_t)}.
\end{equation}
The clipping parameter $\epsilon$ limits the range of policy updates, preventing large deviations from the previous policy.

The advantage $\hat{A}_t$ is estimated via Generalized Advantage Estimation (GAE) to reduce variance:
\begin{equation}
\delta_t = r_t + \gamma V_\phi(s_{t+1}) - V_\phi(s_t), \qquad
\hat{A}_t = \sum_{l=0}^{T-t-1} (\gamma \lambda)^l \, \delta_{t+l},
\end{equation}
where $\gamma$ is the discount factor and $\lambda$ balances bias and variance.

The critic is trained to minimize the squared error between predicted and empirical returns:
\begin{equation}
\mathcal{L}_V(\phi) = \frac{1}{2}\,\mathbb{E}_t\!\left[ \big(V_\phi(s_t)-\hat{R}_t\big)^2 \right],
\end{equation}
where $\hat{R}_t = \sum_{k=t}^T \gamma^{k-t} r_k$ is the empirical return.

Finally, entropy regularization is added to encourage exploration:
\begin{equation}
\mathcal{H}_t = - \sum_a \pi_\theta(a\mid s_t)\,\log \pi_\theta(a\mid s_t).
\end{equation}

The complete loss function for training both policy and value networks is:
\begin{equation}
\mathcal{J}(\theta,\phi) = -\mathcal{L}_{\text{clip}}(\theta) 
+ \beta_V \mathcal{L}_V(\phi) 
- \beta_H \mathbb{E}[\mathcal{H}_t],
\end{equation}
where $\beta_V$ and $\beta_H$ are coefficients for value loss and entropy regularization, respectively. 
By balancing policy improvement, value estimation, and exploration, PPO provides stable and efficient end-to-end training for our SSD-Mamba2 backbone.

\begin{remark}
While PPO provides stable on-policy updates, additional mechanisms are required to ensure robustness in complex locomotion tasks. 
We therefore incorporate \emph{domain randomization}, which perturbs terrain friction, obstacle appearance, and sensor noise during training, and a \emph{curriculum schedule}, which gradually increases obstacle density and terrain complexity. 
These strategies expose the agent to diverse conditions, mitigate overfitting to specific environments, and reduce catastrophic failures such as falls and collisions. 
As a result, SSD-Mamba2 policies not only converge faster but also generalize more reliably to unseen scenarios.
\end{remark}

{
\color{black}
\subsection{Safety Considerations}
\label{sec:safety}

We consider safety during both training and execution, with the goal of improving empirical robustness. 

During training, several mechanisms are employed to mitigate unsafe exploration. Episodes are terminated immediately upon falling or severe collisions, which prevents repeated exposure to hazardous states. The policy outputs bounded joint position targets that are executed through a low-level PD controller, thereby reducing abrupt or unstable actuation. In addition, an obstacle-density curriculum gradually increases task difficulty, enabling the agent to acquire basic locomotion skills before encountering complex or cluttered environments. Domain randomization and visual perturbations further introduce variations in terrain properties, physical parameters, and sensor characteristics, exposing the policy to a wide range of conditions and reducing overfitting.

At execution time, the learned policy operates at a high level by producing joint position targets rather than raw torque commands. This design makes it possible to combine the policy with conventional low-level control and safety mechanisms, such as joint limits, collision monitors, or emergency stop systems. Moreover, the bounded and input-gated state-space updates in the SSD-Mamba2 backbone suppress spurious sensory inputs, contributing to stable behavior under observation uncertainty.
}

\section{Experiments}
\label{sec:fourth}

\subsection{Environment Setup}
All experiments are carried out on a laptop with an Intel Xeon(R) Platinum 8358P CPU (128~cores, 2.6\,GHz base frequency) and a NVIDIA GeForce RTX 3090 GPU (24\,GB, CUDA~12.8). The computing environment is based on Ubuntu~22.04, with physics simulation executed in PyBullet \citep{coumans2021pybullet}. All models are developed in Python~3.8 using PyTorch~2.4.1.

The model is evaluated across three simulated environments that vary in terrain difficulty and obstacle shapes:
\begin{itemize}
  \item \textbf{Thin Obstacle with Goals}: flat terrain populated with numerous thin cuboid obstacles and goals that offering extra rewards .
  \item \textbf{Rugged Terrain with Obstacles and Goals}: uneven, discontinuous ground with a maximum height of 5 cm, same thin obstacles and goals settings as above, requiring careful placement of foothold.
  \item \textbf{Sphere Obstacles with Goals}: flat terrain with sphere-shaped obstacles that has the same shape as the goals.
\end{itemize}

\begin{figure}
\centering
\begin{subfigure}{\linewidth}
\centering
\includegraphics[width=0.49\linewidth]
{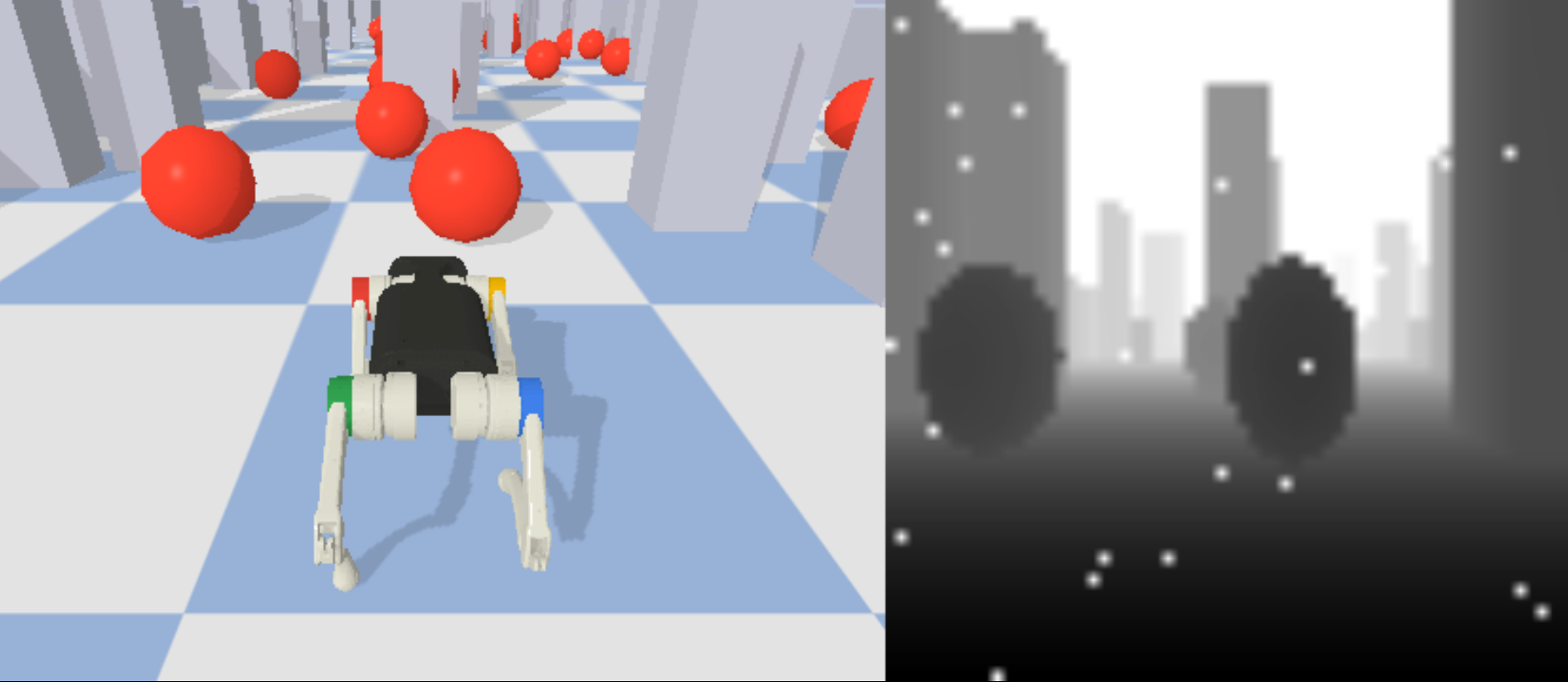}
\hfill
\includegraphics[width=0.49\linewidth]{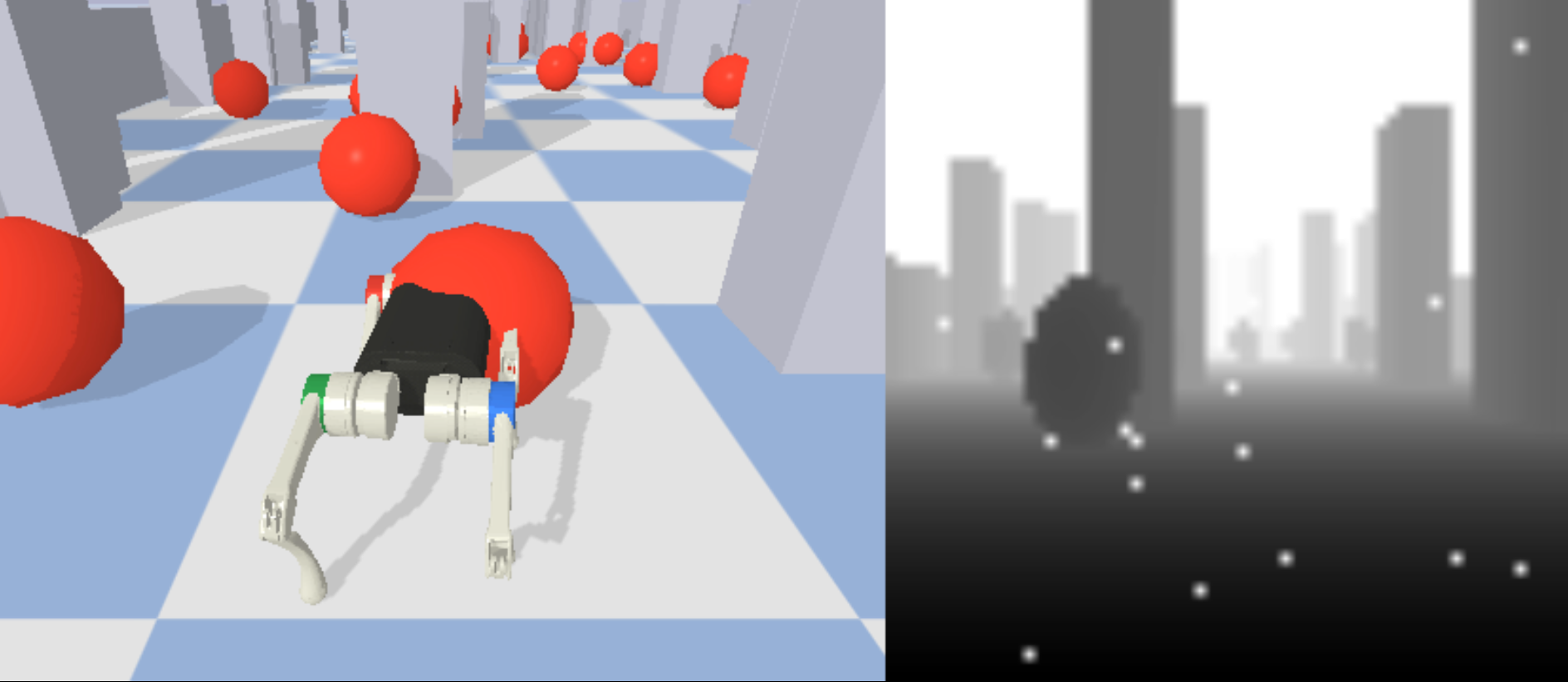} 
    \caption{}
    \label{fig:suba}
\end{subfigure}
\begin{subfigure}{\linewidth}
\centering
\includegraphics[width=0.49\linewidth]
{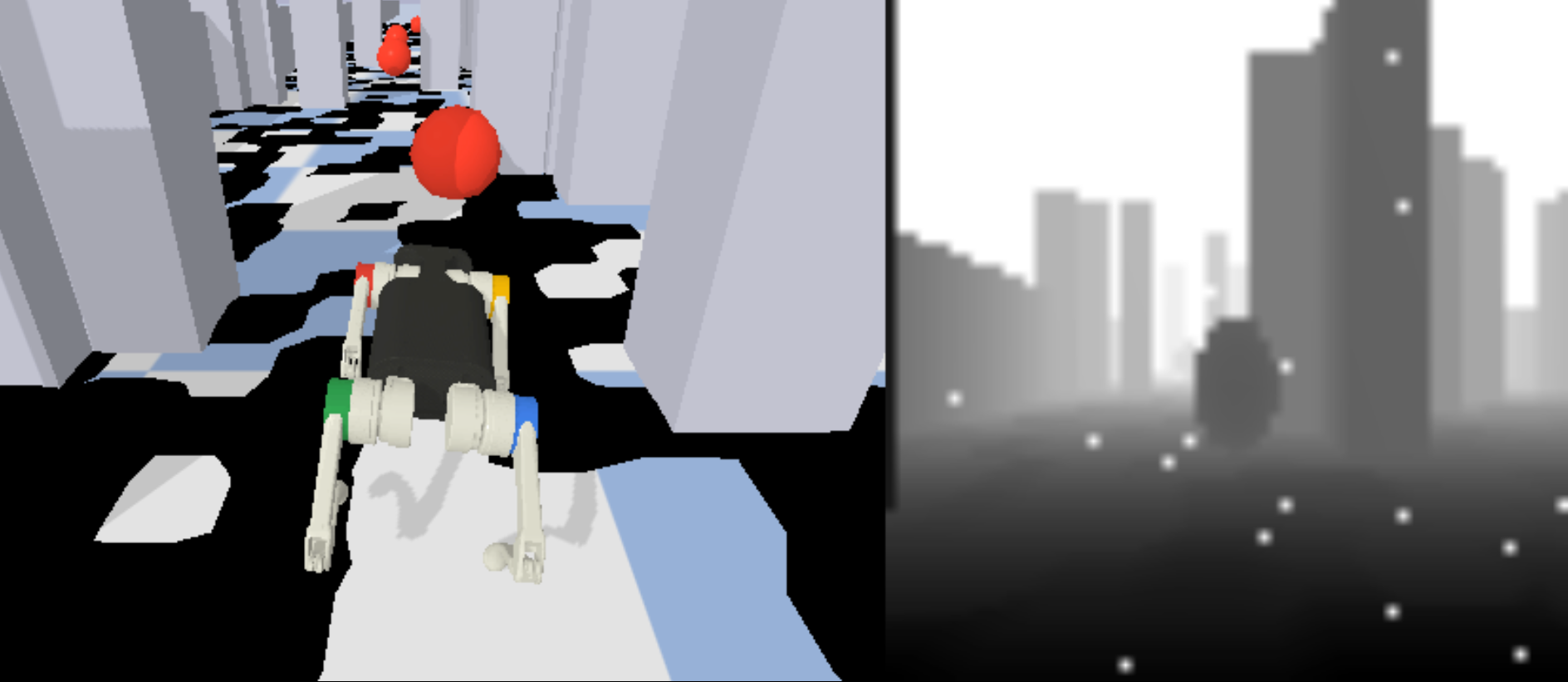}
\hfill
\includegraphics[width=0.49\linewidth]{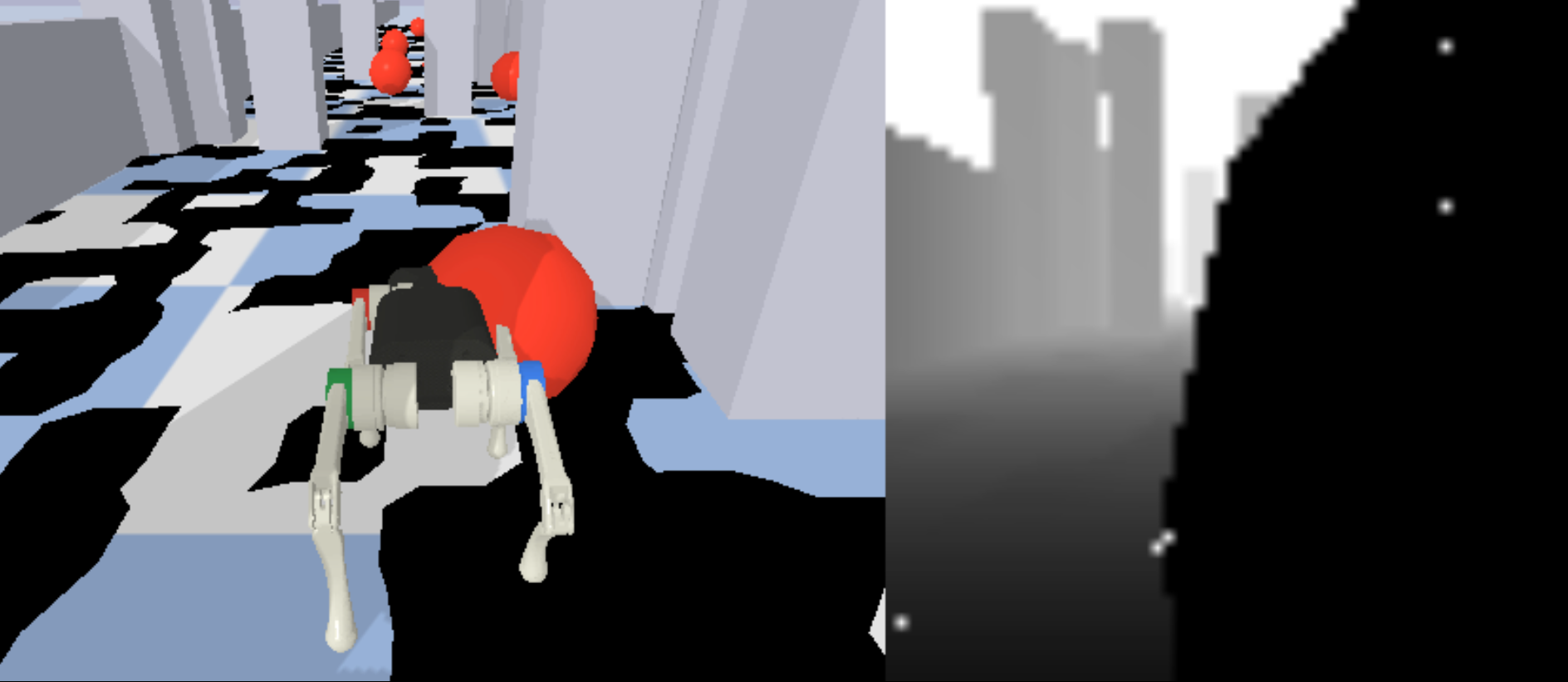} 
    \caption{}
    \label{fig:subb}
\end{subfigure}
\begin{subfigure}{\linewidth}
\centering
\includegraphics[width=0.49\linewidth]
{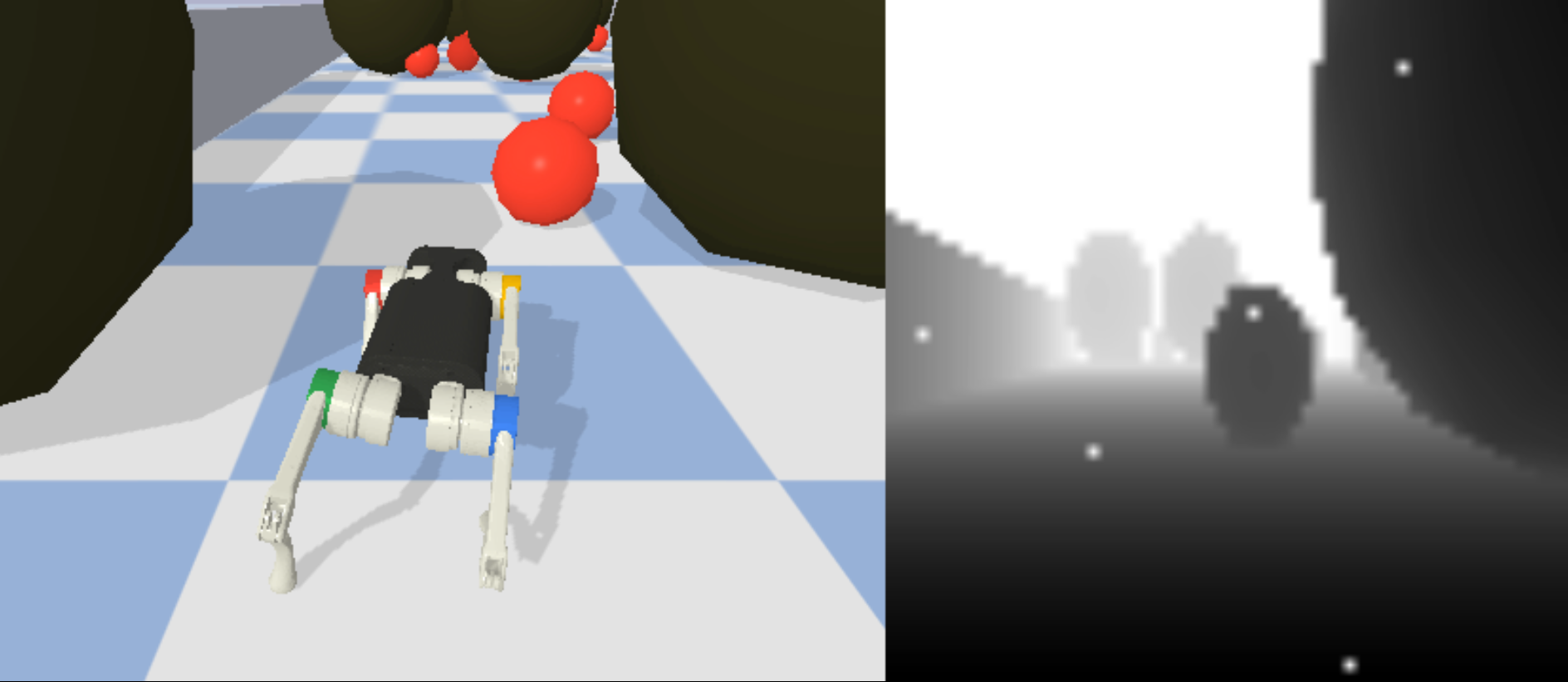}
\hfill
\includegraphics[width=0.49\linewidth]{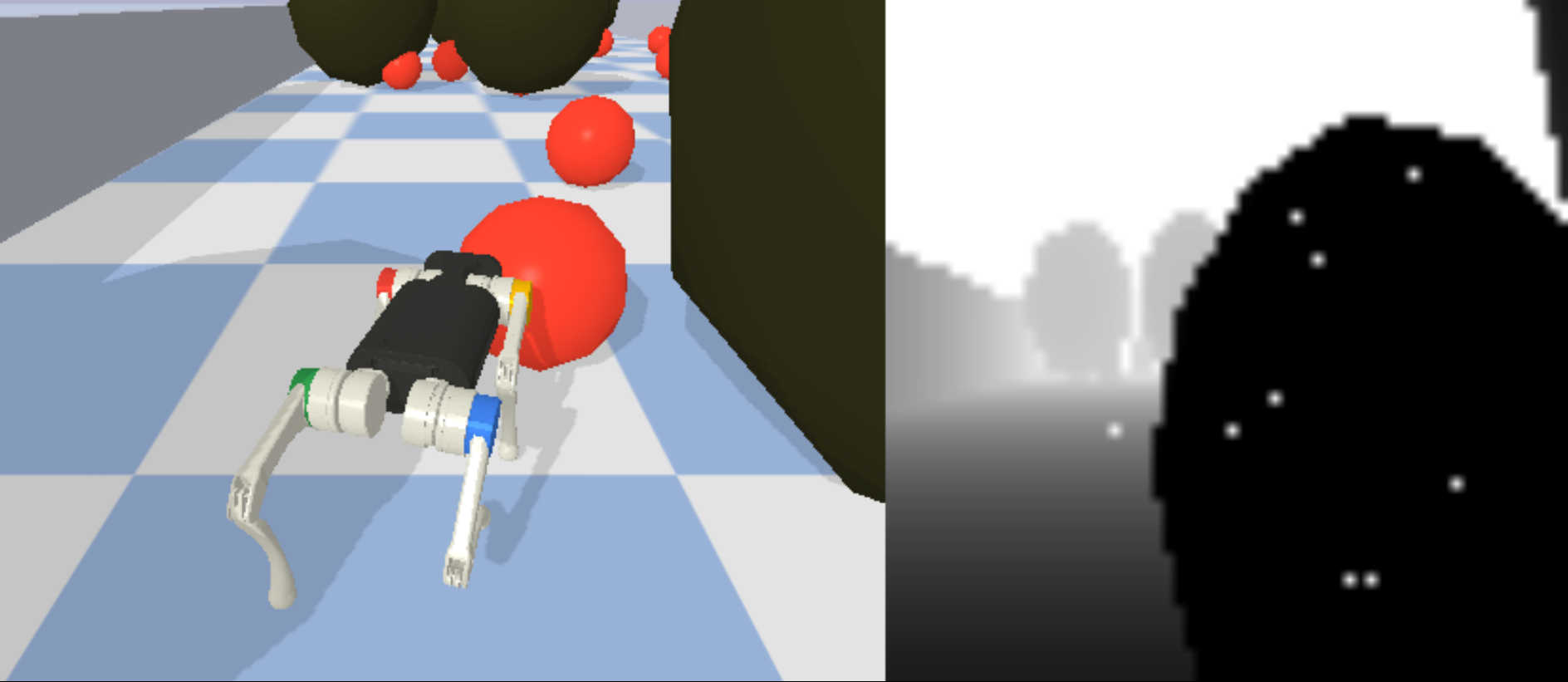} 
    \caption{}
    \label{fig:subc}
\end{subfigure}
\caption{Simulated environments. Panels (a)–(c): (a) Thin Obstacles with Goals; (b) Thin Obstacle with Rugged Terrain; (c) Sphere Obstacles with Goals. Obstacle layouts are randomized at reset; all obstacle positions are static during an episode.}
\label{fig:env_repre}
\end{figure}

Fig. \ref{fig:env_repre} shows representative examples. Unless noted otherwise, obstacle layouts are randomized at episode reset; only \emph{Dynamic Obstacle} updates obstacle positions during an episode.

\subsection{Baseline Methods}
To evaluate the effectiveness of the proposed framework, we compare against the following baselines:

\begin{itemize}
  \item \textbf{Proprio-Only.} A blind agent that relies solely on the 84-dimensional proprioceptive vector, without any exteroceptive input.
  \item \textbf{Transformer Proprio-Vision.} A Transformer-based model that applies self-attention over visual tokens together with a proprioceptive token, representing attention-driven cross-modal fusion \citep{yang2021learning}.
  \item \textbf{Transformer Vision-Only.} A Transformer-based model that uses depth tokens alone as input, following the same architecture as the Transformer Proprio-Vision baseline.
    \item \textbf{Mamba2 Vision-Only.} Using the depth observations alone as input and the same Mamba2-based encoder is used to process the vision tokens.
\end{itemize}

{
\color{black}
To ensure a fair and diagnostic comparison, all baselines are trained with the same PPO algorithm, domain randomization strategy, and curriculum schedule, and are evaluated under identical experimental conditions. Wherever applicable, models share the same vision encoder, proprioceptive encoder, policy head, observation preprocessing pipeline, and training hyperparameters. This controlled setup isolates the effect of multi-modal input and the fusion and temporal modeling backbone, ensuring that observed performance differences can be attributed specifically to these factors rather than to differences in training or implementation.
}

\subsection{Implementation Details}

\paragraph{Model architecture.} 
As illustrated in Fig.~\ref{fig:overall_archi}, the network consists of lightweight encoders, an SSD-Mamba2 fusion backbone, and compact projection heads for policy and value estimation. The detailed architectural settings are summarized in Table~\ref{tab:arch}. 

\begin{table}[t]
\centering
\small
\setlength{\tabcolsep}{6pt}
\caption{Network architecture settings.}
\label{tab:arch}
\begin{tabularx}{\linewidth}{
  @{\hspace{4pt}}
  >{\hspace{2pt}}l
  >{\hspace{2pt}\raggedright\arraybackslash}X
  @{\hspace{4pt}}
}
\toprule
\textbf{Component} & \textbf{Configuration} \\
\midrule
Token width $d$ & 128 \\
Proprio encoder & 2-layer MLP (256, 256), ReLU \\
Visual token projection & CNN patchify $\rightarrow$ linear to width $d$ (=128) \\
Mamba2 backbone & $L_m{=}2$ stacked SSD-Mamba2 layers, residual + LayerNorm \\
Projection head & 2-layer MLP (256, 256), ReLU \\
\bottomrule
\end{tabularx}
\end{table}

\paragraph{Training schema.} 
All policies are trained with PPO using on-policy rollouts of length $T$, minibatch updates over multiple epochs, advantage normalization, and gradient clipping. Terrain and appearance randomization are applied at episode resets, and an obstacle-density curriculum gradually increases task difficulty. The overall training procedure is outlined in Algorithm~\ref{alg:ssdmamba2}. Shared PPO hyperparameters are listed in Table~\ref{tab:rl}.

\begin{algorithm}[t]
\caption{SSD-Mamba2 End-to-End Training with PPO}
\label{alg:ssdmamba2}
\begin{algorithmic}[1]
\Require Policy parameters $\theta$, value parameters $\phi$, fusion module $f_{\text{fuse}}$, horizon $T$, PPO/GAE hyperparameters $(\gamma,\lambda,\epsilon,\beta_V,\beta_H)$, domain randomization ranges $\mathcal{R}$, curriculum scheduler $\mathcal{C}$
\Ensure Trained policy $\pi_\theta$ and value function $V_\phi$

\State Initialize curriculum state $c \gets \mathcal{C}.\textsc{Init}()$

\For{iteration $k = 1,2,\dots$}
  \State Sample environment appearance and obstacle density from $\mathcal{C}$
  \State Initialize replay buffer $\mathcal{D} \gets \emptyset$
  \While{$|\mathcal{D}| < N_{\text{iter}}$}
    \State Sample physics params $\xi \sim \mathcal{R}$; reset environment
    \For{$t=1$ to $T$}
      \State Observe $o_t=\{s^{\text{prop}}_t,I^{\text{depth}}_{t-3:t}\}$ and inject depth noise
      \State Fuse tokens: $h_t \gets f_{\text{fuse}}(o_t)$
      \State Sample action $a_t \sim \pi_\theta(\cdot|h_t)$, step environment, obtain $(r_t,o_{t+1},\text{done})$
      \State Store $(h_t,a_t,r_t,\log\pi_\theta,V_\phi(h_t),\text{done})$ in $\mathcal{D}$
      \If{done} \textbf{break} \EndIf
    \EndFor
  \EndWhile

  \State Compute advantages $\hat{A}_t$ with GAE, normalize, and compute returns $\hat{R}_t$

  \For{epoch $=1$ to $E$}
    \For{minibatch $\mathcal{B}\subset \mathcal{D}$}
      \State Compute $\rho_t$, clipped policy loss $\mathcal{L}_{\text{clip}}$, value loss $\mathcal{L}_V$, and entropy $\mathcal{H}$
      \State Update $(\theta,\phi)$ minimizing $-\mathcal{L}_{\text{clip}} + \beta_V \mathcal{L}_V - \beta_H \mathcal{H}$
    \EndFor
  \EndFor

  \State Advance curriculum: $c \gets \mathcal{C}.\textsc{Advance}(c)$
\EndFor
\end{algorithmic}
\end{algorithm}

\begin{table}[t]
\centering
\caption{PPO hyperparameters.}
\label{tab:rl}
\begin{tabular*}{\linewidth}{
  @{\hspace{4pt}}
  >{\hspace{2pt}}l
  @{\extracolsep{\fill}}
  >{\hspace{2pt}}l
  @{\hspace{4pt}}
}
\toprule
\textbf{Hyperparameter} & \textbf{Value} \\
\midrule
Episode horizon (steps) & 999 \\
Samples per iteration & 16{,}384 \\
Minibatch size & 1{,}024 \\
Optimization epochs per update & 3 \\
Discount factor $\gamma$ & 0.99 \\
GAE parameter $\lambda$ & 0.95 \\
PPO clip parameter $\epsilon$ & 0.2 \\
Entropy coefficient & 0.005 \\
Policy learning rate & $1\times 10^{-4}$ \\
Value learning rate & $1\times 10^{-4}$ \\
Optimizer & Adam \\
Activation function & ReLU \\
\bottomrule
\end{tabular*}
\end{table}

\paragraph{Domain randomization.} 
To improve robustness and transferability, physics parameters are randomized at the beginning of each episode \citep{mehta2020active}. Unless stated otherwise, the sampled values remain fixed throughout the episode. Table~\ref{tab:dranges} lists the randomized ranges.

\begin{table}[t]
\centering
\small
\setlength{\tabcolsep}{6pt}
\caption{Domain randomization ranges.}
\label{tab:dranges}
\begin{tabular*}{\linewidth}{
  @{\hspace{4pt}}
  @{\extracolsep{\fill}}
  >{\hspace{2pt}}l
  >{\hspace{2pt}}l
  >{\hspace{2pt}}l
  @{\hspace{4pt}}
}
\toprule
\textbf{Parameter} & \textbf{Range} & \textbf{Units} \\
\midrule
$K_P$            & $[40,\ 90]$                 & - \\
$K_D$            & $[0.4,\ 0.8]$               & - \\
Link inertia     & $[0.5,\ 1.5]\times$ default & - \\
Lateral friction & $[0.5,\ 1.25]$              & N·s/m \\
Body mass        & $[0.8,\ 1.2]\times$ default & kg \\
Motor friction   & $[0.0,\ 0.05]$              & N·m·s/rad \\
Motor strength   & $[0.8,\ 1.2]\times$ default & N·m \\
Sensor latency   & $[0,\ 0.04]$                & s \\
\bottomrule
\end{tabular*}
\end{table}

\paragraph{Visual perturbations.} 
To simulate sensor artifacts, light-weight noise is injected into depth observations. Specifically, $K\sim\mathcal{U}\{3,\dots,30\}$ pixel locations are selected per frame and their values set to the maximum sensor range, producing salt-like saturation effects. This perturbation is applied consistently across all methods.

\paragraph{Obstacle-density curriculum.} 
In obstacle-rich environments, training begins with sparse obstacle placement and linearly increases density to the target distribution as learning progresses \citep{wang2021survey}. This schedule is applied identically to all compared methods.


{
\color{black}
\paragraph{Parameter selection.}
Parameter selection in this work follows standard practice in reinforcement learning for motion control and is kept fixed across all experiments and baselines. PPO hyperparameters and network configurations adopt commonly used settings and are not tuned per environment or per method. Task-specific parameters, including domain randomization ranges, visual perturbation strength, and obstacle-density curriculum schedules, are selected using simple and interpretable heuristics rather than exhaustive tuning. Domain randomization ranges are defined as relative perturbations around nominal physical parameters to reflect realistic uncertainty. The obstacle-density curriculum follows a monotonic schedule from sparse to dense configurations to gradually increase task difficulty. Visual perturbations are chosen to simulate typical depth-sensor artifacts without overwhelming the underlying observation structure. Overall, this design prioritizes robustness and fair comparison, ensuring that performance gains arise from the proposed framework rather than environment-specific parameter tuning.
}

\subsection{Evaluation Metrics}
The learned policies are assessed using three metrics.  
(i) \textit{Mean episode return}, which summarizes overall task performance.  
(ii) \textit{Distance moved}, defined as the displacement (in meters) achieved along the task-aligned target direction.  
(iii) \textit{Collision times}, measured as the number of collisions recorded until either three evaluation episodes are completed or the robot falls. Collision checks are performed at every control step, and the collision metric is reported only in scenarios containing obstacles and only for episodes in which at least one collision occurs.

\section{Experiment Results}
\label{sec:results}

\subsection{Performance vs.\ Baselines}
\begin{figure}
\centering
\includegraphics[width=\linewidth]{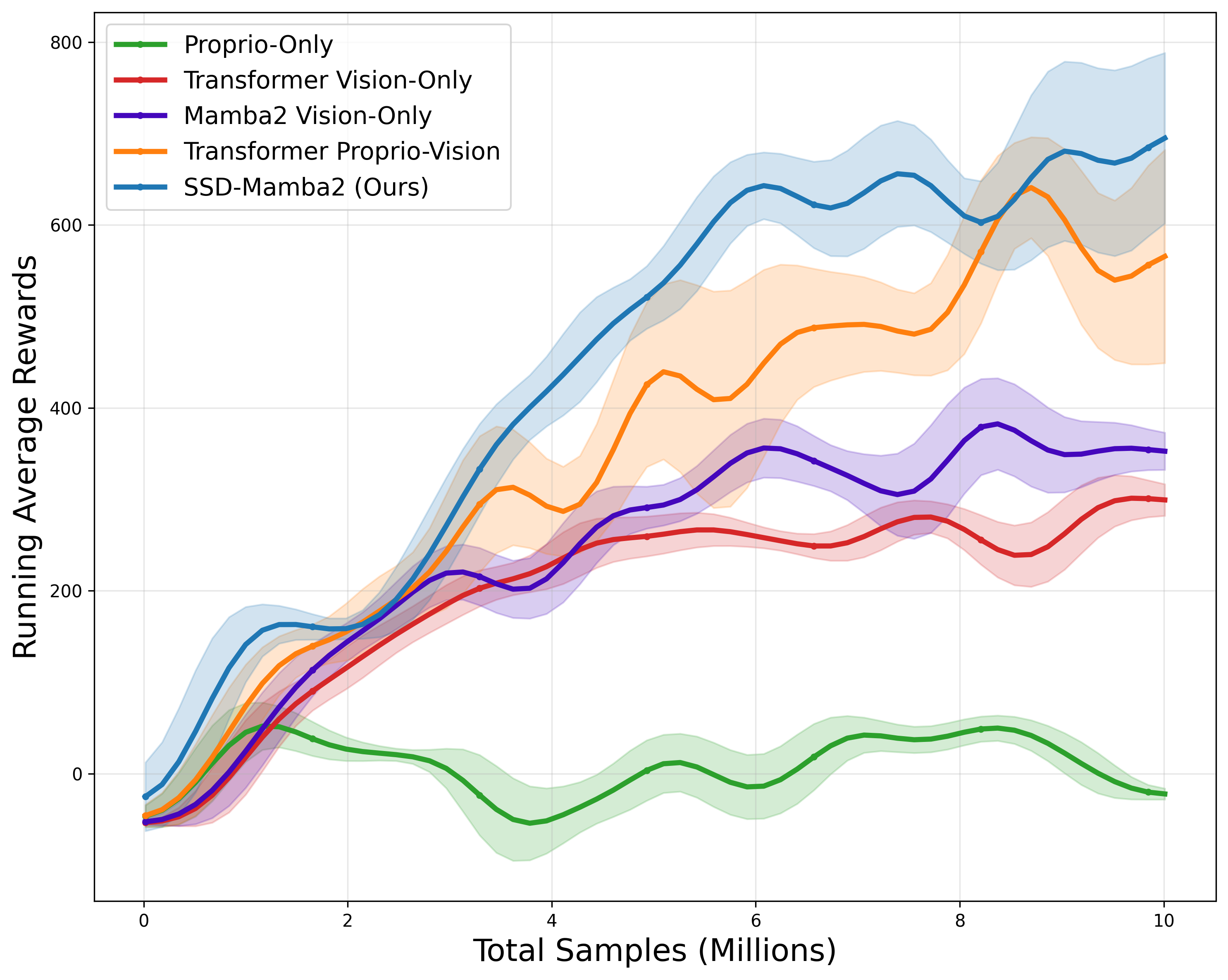}
\caption{Training learning curves on \emph{Thin Obstacle \& Goals}. Solid lines denote mean episode return across seeds; shaded regions indicate $\pm$ one standard deviation.}
\label{fig:learn_curves}
\end{figure}

We benchmark \emph{SSD-Mamba2} against four baselines in the \emph{Thin Obstacle \& Goals} environment. 
Figure~\ref{fig:learn_curves} shows training curves (mean $\pm$ one standard deviation across $10$ seeds), and Table~\ref{tab:performance_comparison} reports final metrics.

\begin{table*}[t]
\centering
\begin{minipage}{0.92\linewidth}
  \centering
  \caption{Performance comparison in the \emph{Thin Obstacle \& Goals} environment. Mean~$\pm$~std over $10$ runs.}
  \label{tab:performance_comparison}
  \small
  \begin{tabular*}{\linewidth}{@{\hspace{4pt}} l @{\extracolsep{\fill}} c c c @{\hspace{4pt}}}
  \toprule
  \textbf{Method} & \textbf{Return} & \textbf{Collision Times} & \textbf{Distance Moved (m)} \\
  \midrule
  Proprio-Only               & 56.34 $\pm$ 33.37  & 571.16 $\pm$ 148.41 & 3.29 $\pm$ 0.97 \\
  Transformer Vision-Only    & 13.44 $\pm$ 35.07  & --                  & 1.27 $\pm$ 0.88 \\
  Mamba2 Vision-Only         & 16.75 $\pm$ 38.22  & --                  & 1.54 $\pm$ 1.04 \\
  Transformer Proprio-Vision & 354.40 $\pm$ 352.59 & 202.47 $\pm$ 131.72 & 7.55 $\pm$ 5.74 \\
  SSD-Mamba2 (Ours)          & \textbf{537.67 $\pm$ 307.79} & \textbf{193.70 $\pm$ 93.61} & \textbf{10.50 $\pm$ 5.36} \\
  \bottomrule
  \end{tabular*}
\end{minipage}
\end{table*}

From Fig.~\ref{fig:learn_curves}, \emph{SSD-Mamba2} shows a steeper early slope, higher asymptotic return, and tighter confidence band than all baselines. 
Numerically (Table~\ref{tab:performance_comparison}), \emph{SSD-Mamba2} improves return over \emph{Transformer Proprio-Vision} from $354.4$ to $537.7$ (+\textbf{51.7}\%), reduces collisions from $202.5$ to $193.7$ (–\textbf{4.3}\%), and increases distance from $7.55$\,m to $10.50$\,m (+\textbf{39.1}\%). 
These results indicate both higher peak performance and more stable optimization.

\textcolor{black}{\subsection{Model Complexity and Resource Usage}}
{
\color{black}
Table~\ref{tab:model_complexity} summarizes the model complexity and resource usage of different policy backbones, providing a quantitative view of their compute--memory trade-offs.

The \emph{Proprio-Only} baseline is the lightest configuration, with $220{,}684$ parameters, $0.18$\,M FLOPs, and minimal memory usage (14.16\,MB allocated), reflecting the absence of visual processing. Introducing vision dominates the arithmetic cost: both \emph{Transformer Vision-Only} and \emph{Mamba2 Vision-Only} incur similar FLOPs (10.60\,M vs.\ 10.57\,M). For multi-modal fusion, \emph{SSD-Mamba2} achieves a favorable balance between model capacity and computational cost. Compared with \emph{Transformer Proprio-Vision}, it reduces the parameter count from $386{,}476$ to $342{,}940$ while maintaining essentially identical FLOPs (11.01\,M vs.\ 11.05\,M). This indicates that the performance improvements reported earlier are not obtained by increasing arithmetic complexity or model size, but instead stem from more effective sequence fusion enabled by the state-space formulation.

In terms of memory usage, \emph{SSD-Mamba2} allocates and reserves more GPU memory (411.39\,MB / 444.00\,MB) than attention-based fusion. This overhead is primarily associated with recurrent state buffers and block-wise scan execution rather than parameter storage. Importantly, this design avoids the quadratic memory growth inherent to attention mechanisms and supports sequential, streaming-friendly computation with near-linear scaling in sequence length.

\begin{table*}[t]
\centering
\caption{\textcolor{black}{Model complexity and resource usage}}
\label{tab:model_complexity}
{\color{black}
\begin{tabular}{lcccccc}
\toprule
\textbf{Method} & \textbf{Total Params} & \textbf{Model Size}  & \textbf{FLOPs}  & \textbf{Memory Allocated} & \textbf{Memory Reserved} \\
& & \textbf{(MB)} & \textbf{(M)} & \textbf{(MB)} & \textbf{(MB)} \\
\midrule
Proprio-Only & 220{,}684 & 0.84 & 0.18 & 14.16 & 28.00 \\
Transformer Vision-Only & 266{,}092 & 1.02 & 10.60 & 139.57 & 150.00 \\
Mamba2 Vision-Only & 222{,}556 & 0.85 & 10.57 & 406.36 & 426.00 \\
Transformer Proprio-Vision & 386{,}476 & 1.47 & 11.05 & 167.56 & 188.00 \\
SSD-Mamba2 (Ours)  & 342{,}940 & 1.31 & 11.01 & 411.39 & 444.00 \\
\bottomrule
\end{tabular}
}
\end{table*}

}

\subsection{Effect of Multi-Modality}

Single-modality agents perform poorly. 
\emph{Proprio-Only} achieves a return of $56.3$, with $571.2$ collisions and only $3.29$\,m traveled. 
Vision-only agents barely progress: \emph{Transformer Vision-Only} reaches a return of $13.4$ with $1.27$\,m distance, while \emph{Mamba2 Vision-Only} yields $16.8$ return and $1.54$\,m distance.

Adding vision to proprioception provides substantial gains. 
\emph{Transformer Proprio-Vision} improves over \emph{Proprio-Only} by raising return from $56.3$ to $354.4$ (+\textbf{529}\%), reducing collisions from $571.2$ to $202.5$ (–\textbf{64.5}\%), and increasing distance from $3.29$\,m to $7.55$\,m (+\textbf{129.5}\%). 
It also outperforms \emph{Transformer Vision-Only}, achieving a return more than \textbf{25.4}\,$\times$ higher ($354.4$ vs.\ $13.4$) and a distance nearly \textbf{6}\,$\times$ longer ($7.55$\,m vs.\ $1.27$\,m), while providing meaningful collision reduction.

Our proposed \emph{SSD-Mamba2} further amplifies these benefits. 
Compared with \emph{Mamba2 Vision-Only}, it delivers over \textbf{30}\,$\times$ higher return ($537.7$ vs.\ $16.8$) and about \textbf{6.8}\,$\times$ longer distance ($10.50$\,m vs.\ $1.54$\,m). 
Relative to \emph{Proprio-Only}, it boosts return nearly \textbf{9.5}\,$\times$ ($537.7$ vs.\ $56.3$) while cutting collisions by \textbf{66\%} ($193.7$ vs.\ $571.2$). These results confirm that multi-modality is essential for foresightful quadrupedal locomotion control.

\subsection{Effect of the Fusion Backbone}

Among vision-only agents, \emph{Mamba2 Vision-Only} provides modest gains over \emph{Transformer Vision-Only}, improving return from $13.4$ to $16.8$ (+\textbf{25.4}\%) and distance from $1.27$\,m to $1.54$\,m (+\textbf{21.3}\%). However, both remain far from the performance of multi-modal agents. 

For multi-modal inputs, \emph{SSD-Mamba2} outperforms \emph{Transformer Proprio-Vision}, increasing return from $354.4$ to $537.7$ (+\textbf{51.7}\%), raising distance from $7.55$\,m to $10.50$\,m (+\textbf{39.1}\%), and slightly lowering collisions from $202.5$ to $193.7$ (–\textbf{4.3}\%).

Overall, the backbone choice becomes decisive once multi-modal inputs are available: SSD-Mamba2 delivers more effective fusion and superior control outcomes than attention-based fusion under the same training protocol.

\subsection{Mechanism Visualization: Selectivity and Gating}
To gain insight into how \emph{SSD-Mamba2} processes multimodal inputs, we visualize the internal \emph{selectivity} and \emph{gating} patterns for two representative layers.

\begin{figure*}
\centering
\includegraphics[width=0.95\linewidth]{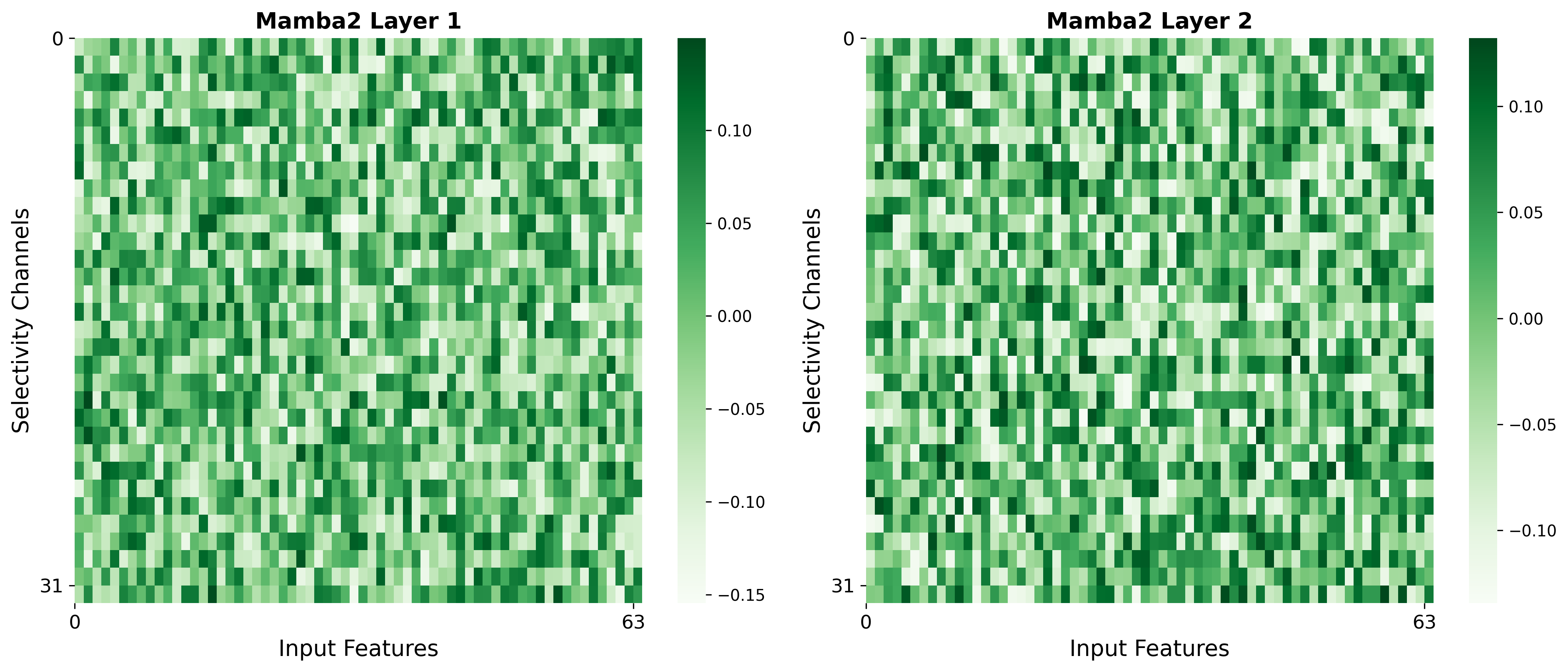}
\caption{Selectivity heatmaps of \emph{SSD-Mamba2} (Layer~1 and Layer~2). The \emph{x}-axis denotes input features and the \emph{y}-axis denotes selectivity channels. Brighter cells indicate stronger retention, while darker cells indicate suppression.}
\label{fig:select_map}
\end{figure*}

\textbf{Selectivity.}  
Figure~\ref{fig:select_map} shows that both layers exhibit structured, non-uniform selectivity rather than uniform weighting across tokens. This indicates that the model actively learns which input channels to retain and which to suppress. Specifically, proprioceptive channels that encode stability cues consistently receive stronger weights, while subsets of depth tokens corresponding to obstacles or terrain irregularities are selectively retained. This aligns with our quantitative results: \emph{SSD-Mamba2} achieves substantially fewer collisions and longer travel distances compared with baselines, suggesting that selective filtering of proprioceptive and hazard-related cues contributes directly to safer and more foresightful locomotion.

\begin{figure*}
\centering
\includegraphics[width=0.95\linewidth]{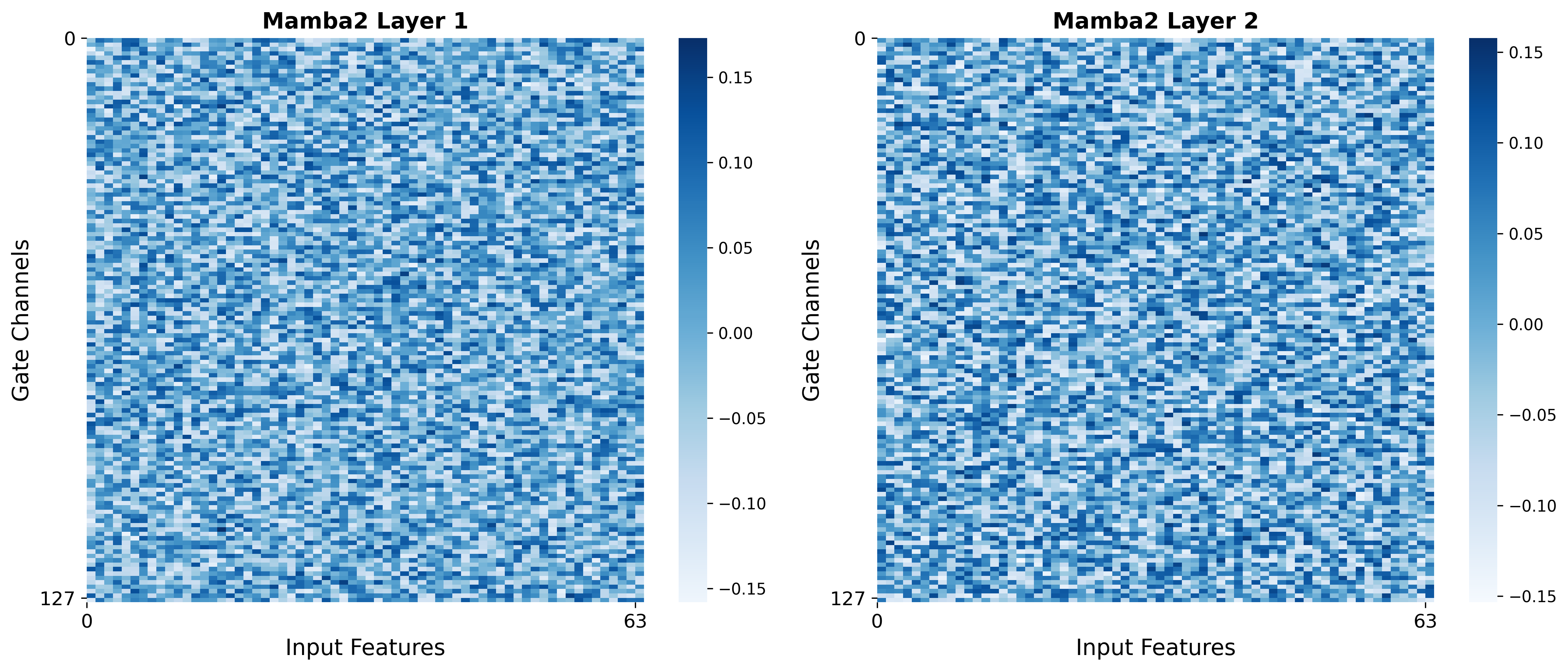}
\caption{Gating heatmaps of \emph{SSD-Mamba2} (Layer~1 and Layer~2). The \emph{x}-axis denotes input features and the \emph{y}-axis denotes gating channels. Brighter cells indicate stronger pass-through to the output, while darker cells indicate suppression.}
\label{fig:gating_map}
\end{figure*}

\textbf{Gating.}  
The gating mechanism, shown in Fig.~\ref{fig:gating_map}, regulates which processed features are propagated to the next layer after state-space scanning. We observe stronger gating for features associated with obstacle proximity and base stabilization, while redundant or noisy depth tokens are consistently suppressed. This selective pass-through acts as an information bottleneck that emphasizes task-relevant features and discards irrelevant signals. Such behavior explains two key empirical findings: (i) the faster convergence of \emph{SSD-Mamba2} in Fig.~\ref{fig:learn_curves}, since the model reduces noise early during training; and (ii) the higher asymptotic return, as policies are guided by features most critical for progress and safety.

Together, these visualizations provide a mechanistic interpretation: by selectively retaining proprioceptive stability cues, emphasizing obstacle-related depth information, and gating outputs to prioritize task-relevant features, the model achieves both efficient learning and robust control.

\subsection{Generalization to Unseen Conditions}
A central challenge for quadrupedal controllers is whether policies trained in one environment can transfer to unseen conditions without fine-tuning. 
We therefore train all agents in \emph{Thin Obstacle \& Goals} and evaluate them zero-shot in two novel scenarios: \emph{Rugged Terrain with Thin Obstacles \& Goals} and \emph{Sphere Obstacles \& Goals}. 
These settings introduce irregular surfaces and altered obstacle geometry, testing the robustness and foresight of the learned policies.

\begin{table*}
\centering
\begin{minipage}{0.92\linewidth}
  \centering
  \caption{Zero-shot performance on \emph{Rugged Terrain with Thin Obstacles \& Goals}. Mean $\pm$ std over 10 runs.}
  \label{tab:generalization_rugged}
  \small
  \begin{tabular*}{\linewidth}{
    @{\hspace{4pt}} l
    @{\extracolsep{\fill}} c c c
    @{\hspace{4pt}}
  }
  \toprule
  \textbf{Method} & \textbf{Return} & \textbf{Collision Times} & \textbf{Distance Moved (m)} \\
  \midrule
  Proprio-Only               & 70.24 $\pm$ 41.11  & 735.37 $\pm$ 120.63 & 3.44 $\pm$ 1.26 \\
  Transformer Vision-Only    & 52.29 $\pm$ 54.78  & --                  & 2.34 $\pm$ 1.48 \\
  Mamba2 Vision-Only         & 12.07 $\pm$ 62.11  & --                  & 1.68 $\pm$ 1.65 \\
  Transformer Proprio-Vision & 432.32 $\pm$ 182.28 & \textbf{560.47 $\pm$ 152.65} & 9.22 $\pm$ 3.01 \\
  SSD-Mamba2 (Ours)          & \textbf{601.62 $\pm$ 289.57} & 638.93 $\pm$ 118.38 & \textbf{11.31 $\pm$ 4.94} \\
  \bottomrule
  \end{tabular*}
\end{minipage}
\end{table*}

\begin{table*}
\centering
\begin{minipage}{0.92\linewidth}
  \centering
  \caption{Zero-shot performance on \emph{Sphere Obstacles \& Goals}. Mean $\pm$ std over 10 runs.}
  \label{tab:generalization_dynamic}
  \small
  \begin{tabular*}{\linewidth}{
    @{\hspace{4pt}} l
    @{\extracolsep{\fill}} c c c
    @{\hspace{4pt}}
  }
  \toprule
  \textbf{Method} & \textbf{Return} & \textbf{Collision Times} & \textbf{Distance Moved (m)} \\
  \midrule
  Proprio-Only               & 85.48 $\pm$ 92.37  & 482.37 $\pm$ 136.02 & 5.50 $\pm$ 2.36 \\
  Transformer Vision-Only    & 41.02 $\pm$ 36.82  & --                  & 2.15 $\pm$ 0.91 \\
  Mamba2 Vision-Only          & 82.74 $\pm$ 45.02  & --                  & 3.87 $\pm$ 0.84 \\
  Transformer Proprio-Vision & 467.56 $\pm$ 230.68 & 163.17 $\pm$ 77.44  & 10.26 $\pm$ 4.54 \\
  SSD-Mamba2 (Ours)          & \textbf{608.47 $\pm$ 344.02} & \textbf{158.77 $\pm$ 117.62} & \textbf{12.24 $\pm$ 7.04} \\
  \bottomrule
  \end{tabular*}
\end{minipage}
\end{table*}

\textbf{Rugged terrain} (Table~\ref{tab:generalization_rugged}): 
On uneven surfaces, proprioception alone is insufficient, yielding only $70.2$ return and $3.44$\,m distance. 
Adding vision improves foresight: \emph{Transformer Proprio-Vision} reaches $432.3$ return and $9.22$\,m distance, demonstrating the benefit of multi-modal fusion. 
\emph{SSD-Mamba2} further improves return to $601.6$ (+\textbf{39.2}\%) and distance to $11.31$\,m (+\textbf{22.7}\%) over \emph{Transformer Proprio-Vision}, while keeping collisions competitive (638.9 vs.\ 560.5). 
The slightly higher collision rate (+\textbf{14.0}\%) suggests that our policy explores more aggressively to achieve longer progress on rugged ground. 
Overall, \emph{SSD-Mamba2} attains the best performance in return and distance, underscoring the resilience of selective state-space fusion.

\textbf{Sphere obstacles} (Table~\ref{tab:generalization_dynamic}): 
With spherical hazards, single-modality agents again fail to progress meaningfully, moving less than $6$\,m on average. 
\emph{Transformer Proprio-Vision} improves substantially ($467.6$ return, $10.26$\,m distance, $163.2$ collisions). 
\emph{SSD-Mamba2} surpasses it with $608.5$ return (+\textbf{30.1}\%), $12.24$\,m distance (+\textbf{19.3}\%), and slightly fewer collisions (158.8, –\textbf{2.7}\%). 
This highlights that our backbone not only achieves higher asymptotic performance but also generalizes better to novel obstacle geometries.

Across both unseen settings, \emph{SSD-Mamba2} consistently outperforms strong baselines. 
On rugged terrain, it prioritizes progress while maintaining competitive safety; in dynamic obstacle settings, it achieves both higher returns and fewer collisions. 
Together with the in-distribution experiments and mechanism visualizations, these findings demonstrate that selective state-space fusion enables policies that are robust, foresightful, and transferable beyond their training environments.

{
\color{black}
\subsection{Impact of Input Frame Length}
To examine whether longer instantaneous visual context provides additional benefits, we perform an ablation study by varying the number of stacked depth frames from 4 to 8 and 16 while keeping all other training settings unchanged. This isolates the effect of input frame length from the model’s internal temporal aggregation mechanism.

The results are summarized in Table~\ref{tab:performance_comparison}. In the \emph{Thin Obstacle \& Goals} environment, stacking 4 frames yields the best overall performance, achieving the highest return (537.67), the fewest collisions (193.70), and the longest travel distance (10.50\,m). Increasing the input length to 8 or 16 frames consistently degrades performance across all metrics, indicating that excessive visual history introduces noise rather than useful foresight in this relatively structured setting.

\begin{table*}[t]
\centering
\caption{\textcolor{black}{Performance comparison under different input frame lengths. Mean~$\pm$~std over $10$ runs.}}
\label{tab:performance_comparison}
\small
{\color{black}
\begin{tabular*}{\linewidth}{@{\hspace{4pt}} c @{\extracolsep{\fill}} c c c @{\hspace{4pt}}}
\toprule
\textbf{Input Frame Length} & \textbf{Return} & \textbf{Collision Times} & \textbf{Distance Moved (m)} \\
\midrule

\multicolumn{4}{c}{\textbf{Thin Obstacle \& Goals Environment}} \\
\midrule
4  & \textbf{537.67 $\pm$ 307.79} & \textbf{193.70 $\pm$ 93.61} & \textbf{10.50 $\pm$ 5.36} \\
8  & 492.41 $\pm$ 258.61          & 222.53 $\pm$ 96.65          & 9.77 $\pm$ 3.37 \\
16 & 466.72 $\pm$ 329.21          & 289.33 $\pm$ 145.84         & 8.09 $\pm$ 4.19 \\
\midrule

\multicolumn{4}{c}{\textbf{Rugged Terrain with Thin Obstacles \& Goals Environment}} \\
\midrule
4  & \textbf{601.62 $\pm$ 289.57} & 638.93 $\pm$ 118.38 & \textbf{11.31 $\pm$ 4.94} \\
8  & 595.79 $\pm$ 324.37          & \textbf{583.77 $\pm$ 79.23} & 11.02 $\pm$ 5.29 \\
16 & 391.48 $\pm$ 237.22          & 658.70 $\pm$ 123.47         & 7.69 $\pm$ 3.79 \\
\midrule

\multicolumn{4}{c}{\textbf{Sphere Obstacles \& Goals Environment}} \\
\midrule
4  & 608.47 $\pm$ 344.02          & 158.77 $\pm$ 117.62         & \textbf{12.24 $\pm$ 7.04} \\
8  & 605.49 $\pm$ 420.51          & \textbf{142.10 $\pm$ 59.34} & 11.62 $\pm$ 6.94 \\
16 & \textbf{609.38 $\pm$ 418.25} & 143.87 $\pm$ 93.48          & 11.64 $\pm$ 7.68 \\
\bottomrule
\end{tabular*}
}
\end{table*}

A similar trend is observed in the \emph{Rugged Terrain with Thin Obstacles \& Goals} environment. While stacking 8 frames slightly reduces collision counts (583.77 vs.\ 638.93), it does so at the cost of lower return and shorter travel distance. In contrast, 4-frame input achieves the highest return (601.62) and the longest distance traveled (11.31\,m), suggesting a better balance between progress and safety. Stacking 16 frames leads to a pronounced performance drop, with both return and distance substantially reduced, highlighting the difficulty of exploiting long raw input sequences in uneven terrain.

In the \emph{Sphere Obstacles \& Goals} environment, the optimal number of input frames varies across metrics: 16 frames marginally improve return, 8 frames minimize collisions, and 4 frames maximize travel distance. This divergence further indicates that longer input sequences do not uniformly enhance control performance and may bias the policy toward specific objectives at the expense of others.

Overall, stacking 4 frames provides a favorable trade-off between perceptual context, optimization stability, and real-time responsiveness. Combined with SSD-Mamba2’s selective state-space temporal aggregation, this compact input window enables effective long-horizon reasoning while avoiding the redundancy and instability associated with longer raw input sequences.

}
\section{Conclusion}
\label{sec:fifth}

This work introduces \emph{SSD-Mamba2}, a vision-driven cross-modal reinforcement learning framework for motion control. By encoding proprioceptive states with an MLP, patchifying depth images using a lightweight CNN, and fusing these representations through stacked SSD-Mamba2 layers, the method leverages state-space duality to achieve efficient long-horizon modeling with near-linear complexity. Coupled with PPO and a compact reward design under domain randomization and curriculum scheduling, the framework enables stable and scalable end-to-end training.

Extensive experiments demonstrate that \emph{SSD-Mamba2} consistently outperforms proprioception-only, vision-only, and Transformer-based baselines in terms of return, safety (fewer collisions and falls), and distance traveled. \textcolor{black}{Learning curves and variance across random seeds further demonstrate superior convergence speed, sample efficiency, and robustness.} Ablation studies highlight the critical role of cross-modal fusion and validate the effectiveness of SSD-Mamba2 as a fusion backbone for safety-critical locomotion.

{
\color{black}
While the present study focuses on empirical robustness under high-fidelity simulation, the deployment of the learned policies on physical quadrupedal platforms remains an important direction for future work. Future efforts will extend the proposed framework to real-world hardware, with particular emphasis on sim-to-real transfer through domain randomization and sensor-noise modelling, real-time inference latency under onboard computation constraints, and systematic safety validation in field conditions. In addition, the integration of explicit safety constraints and safety-aware RL mechanisms will be investigated to further enhance policy reliability during both training and execution.} Beyond locomotion, the proposed approach may also benefit broader motion-control domains such as mobile navigation and human–robot interaction, where efficient fusion of multimodal signals is crucial.

\printcredits

\section*{Declaration of competing interest}
The authors declare that they have no known competing financial interests or personal relationships that could have appeared to influence the work reported in this paper.

\section*{Acknowledgment}
This research was funded by the Graduate Innovation Fund under Grant 101832020CX130.

\section*{Data availability}
Data will be made available on request.

\bibliographystyle{model1-num-names}

\bibliography{cas-refs}


\end{document}